\documentclass{article}

\usepackage{spconf}
\usepackage{graphicx}
\usepackage{amssymb}
\usepackage{epstopdf}
\usepackage[ruled,norelsize]{algorithm2e}
\makeatletter
\newcommand{\removelatexerror}{\let\@latex@error\@gobble}
\makeatother
\usepackage{subfig}
\usepackage{pseudocode}
\usepackage{fancybox}
\usepackage{multicol}
\usepackage{verbatim}
\usepackage{amsthm}
\usepackage{amsmath}
\usepackage{color}
\usepackage{bm}
\usepackage{cite}
\usepackage{multirow}
\usepackage{tikz}
\usepackage{hyperref}
\usepackage{booktabs}

\newcommand{\mbf}[1]{{\mathbf #1}}

\title{Novel Structured Low-rank algorithm to recover spatially smooth exponential image time series}
\name{Arvind Balachandrasekaran and Mathews Jacob \thanks{This work is supported by NIH 1R01EB019961-01A1}}%
\address{Department of Electrical and Computer Engineering, University of Iowa, IA, USA.  \\}

\begin{document}
%
\maketitle
%

\begin{abstract}
We propose a structured low rank matrix completion algorithm to recover a time series of images consisting of linear combination of exponential parameters at every pixel, from undersampled Fourier measurements. The spatial smoothness of these parameters is exploited along with the exponential structure of the time series at every pixel, to derive an annihilation relation in the $k-t$ domain. This annihilation relation translates into a structured low rank matrix formed from the $k-t$ samples. We demonstrate the algorithm in the parameter mapping setting and show significant improvement over state of the art methods. 
\end{abstract}
 \vspace{-1em}
\section{introduction}

The recovery of linear combination of damped exponentials from few of their measurements is a classical problem in signal processing, starting with the seminal work of Prony \cite{stoica_spec}. Early work, which focused on uniform sampling, relied on the existence of a filter that annihilates the uniform measurements. Recently, several researchers have extended this idea to the non-uniform sampling setting, which is a more efficient measurement strategy. These methods pose the estimation as the completion of a structured matrix (e.g Hankel or Toeplitz), whose entries are the signal samples, from few of its measurements. Specifically, the annihilation filters are the null space vectors of the Toeplitz matrix, or equivalently the matrix is low-rank. 

The above exponential estimation problem is of high significance in several medical imaging applications, including MR spectroscopic imaging and MR parameter mapping. These methods perform a pixel-by-pixel estimation of the exponential parameters (e.g. frequencies, decay rate, amplitudes) from an image time series. The parameters provide valuable clues about the abnormal metabolic activity and tissue micro-structural changes, which are early markers for neurological disorders. The parameters often vary smoothly in space since they depend on the underlying tissue microstructure. Unfortunately, the acquisition of the image time series at high spatial resolution results in prohibitively long scan time. A common approach to accelerate the acquisition is to acquire under sampled data, followed by reconstruction using low rank and sparsity priors.

We introduce a novel formulation that directly exploits the exponential structure of the time series and the spatial smoothness of the exponential parameters. Specifically, we couple the time series estimation problems at all the pixels into a single annihilation relation, involving a spatially smooth annihilation filter. This approach exploits the spatial smoothness naturally; we do not require additional spatial regularization priors. The annihilation relations translate to a low-rank penalty on a Toeplitz matrix, whose entries are the $k-t$ space samples of the time series. 

A challenge with the above formulation is the large size of the Toeplitz matrix. The direct implementation of the structured low-rank problem will be prohibitive for high resolution applications. We generalize our generic iteratively reweighted annihilating filter (GIRAF) framework, originally introduced for piecewise smooth images \cite{GIRAF}, to accelerate the computations. In particular, we use an iterative reweighted formulation, where we exploit the Toeplitz structure of the matrix to avoid its direct computation and storage. The GIRAF scheme approximates the linear convolutions resulting from the Toeplitz multiplications with circular convolutions, which allowed their evaluation using Fast Fourier Transform (FFT). These approximations were enabled by the fast decay of the image Fourier coefficients towards the edges of the observed region. Unfortunately, this approach breaks down in our setting since we rely on annihilation relations in $k-t$ space, where the signal has considerable intensity at the first image. Hence, we modify the GIRAF steps using a hybrid circular-linear convolution strategy. These generalizations provide a fast algorithm that can be readily applied to large-scale exponential estimation problems. Our validations using MRI data clearly demonstrate the benefits offered by the proposed algorithm, which can simultaneously exploit the exponential structure and the spatial smoothness of the parameters. 

This work has similarities to recent works on structured low-rank priors for the estimation of piecewise smooth signals and exponential signals. For example,  \cite{ALOHAPM} exploited the smoothness of the $k_{x}-t$ signal using a low-rank prior on a wavelet transform weighted Hankel matrix. This work does not exploit the exponential structure of the signal. In addition, the recovery of each $k_{y}-t$ slice is decoupled into a separate problem to keep the computational complexity reasonable \cite{ALOHAPM}. This decoupled approach is less constrained and cannot handle efficient 2-D sampling patterns; Cartesian patterns with fully sampled $k_{x}$ lines were considered in \cite{ALOHAPM}. In \cite{MORASA}, the linear predictability of the time series at each pixel is exploited individually. Since this decoupled strategy is less constrained, the authors rely on additional low-rank and wavelet sparsity regularization penalties; the optimization of several regularization parameters is also challenging. While the piecewise smoothness and sparsity has also been exploited by several researchers using the structured low-rank setting \cite{loraks}\cite{ALOHA},\cite{gregsiam} they do not exploit the exponential signal structure. 

\vspace{-1em}
\section{Problem formulation}

We focus on the recovery of a 3-D volume $\rho$, consisting of a linear combination of damped exponentials at every pixel, from noisy and undersampled measurements denoted by $\mbf b$. We will now introduce the annihilation relations, and the resulting structured low-rank priors. 

\vspace{-1em}
\subsection{Annihilation of spatially smooth exponentials}
We model the temporal signal at the spatial location $\mathbf r = (x,y)\in \mathbb Z^2$ as :
\begin{equation}
\label{eq:signal model}
\rho[\mathbf{r},n]= \sum_{i=1}^{L}\alpha_{i}(\mathbf{r})~\beta_{i}(\mathbf{r})^{n}.
\end{equation}
Note that the temporal signal $\rho[\mathbf{r},n]$ is a linear combination of $L$ exponentials, $\alpha_{i}(\mathbf{r}) \in \mathbb C$ are the amplitudes and $\beta_{i}(\mathbf r) \in \mathbb C$ are the exponential parameters that are dependent on the physiological process. For example, in $T_2$ mapping applications, the exponential parameters $\beta_i = \exp\left(\frac{-\Delta T}{T_{2,i}(\mathbf r)}\right)$, where $\Delta T$ is the time between two image frames and $T_{2,i}$ is the relaxation parameter of the $i^{\rm th}$ tissue component (e.g. gray matter, CSF, white matter) at the voxel indexed by $\mbf r$.

It is well known that the exponential signal in \eqref{eq:signal model}, corresponding to a specific spatial location $\mathbf r$, can be annihilated by a filter $h[\mathbf{r},n]$ \cite{stoica_spec}:
\begin{equation}
\label{eq:annihil_1d}
\sum_{k \in \theta} \rho[\mathbf r,n-k]~h[\mathbf r,k] = 0 =  \mu[\mathbf r,n],~~ \forall \mbf r.
\end{equation}
where (\ref{eq:annihil_1d}) represents a 1-D convolution and $\theta$ is a valid index set. The coefficients of $h[\mathbf r,n]$ at $\mathbf r$ are specified by $h[\mathbf{r},z]=\prod_{i=1}^{L}(1-\beta_{i}(\mathbf{r})z^{-1})$. Computing the 2-D Fourier transform (along the spatial dimension, denoted by $\mathbf r$) of \eqref{eq:annihil_1d}, we obtain
\begin{equation}
\label{eq:annihil_3d}
 \hat\rho[\mathbf{k},n] \otimes c[\mathbf{k},n] = 0.
\end{equation}
Here,  $\hat \rho[\mathbf k,n]$ are the spatial (2-D) Fourier coefficients of the measurements $\rho[{\mathbf{r}},n]$ and  $\otimes$ denotes 3D convolution. Similarly, $c[\mathbf{k},n]$ denotes the spatial (2-D) Fourier coefficients of $h[{\mathbf{r}},n]$. We assume that the parameters $\beta_i(\mathbf r)$ describing the physiological process are bandlimited functions of the spatial variable $\mathbf r$; this implies that $c[\mathbf{k},n]$ is a 3D FIR filter. In particular,  the extent of the filter $c[\mathbf{k},n]$ in the spatial frequency ($\mathbf k$) dimension controls the spatial smoothness of the filter $h[{\mathbf r},n]$, while its extent in the temporal dimension ($n$) depends on the number of exponentials in the model \eqref{eq:signal model}. We express (\ref{eq:annihil_3d}) in a matrix form as
\begin{equation}
\label{eq:anni3d_matrix}
\mathcal{T}(\hat{\rho})~\mathbf {c} = \mathbf{Q} \mathbf c = 0
\end{equation}
where $\mathcal{T}$ is a linear operator that maps a 3D volume $x$ to a lifted matrix $\mathcal{T}(x) \in \mathbb{C}^{M \times N}$. Here the lifted matrix has a multi-fold Toeplitz structure, such that $\mathbf Q \mathbf c$ corresponds to the 3-D convolution between $\hat \rho[\mathbf k,n]$ and the FIR filter $c[\mathbf k,n]$. The number of columns of the matrix $\mathcal T(\hat{\rho})$ is equal to the assumed support of the filter. Likewise, if the measurements $ \hat\rho[\mathbf{k},n] $ are restricted to a cube shaped volume in $(\mathbf k,n)$, the rows correspond to the valid convolutions. 

In practice, the support of the filter $\mathbf c[\mathbf k,n]$ is unknown. Let us assume that (\ref{eq:annihil_3d}) is satisfied by a minimal filter $\mathbf c[\mathbf k,n]$ of support $\Delta \subset \mathbb Z^3$. In this case, we observe that
\begin{equation}
 \hat\rho[\mathbf{k},n] \otimes d[\mathbf{k},n] = 0,
\end{equation}
where $d = c \otimes e$ and $e[\mathbf{k},n]$ is any FIR filter. Note that the support of $d$, denoted by $\Theta$ is larger than $\Delta$; i.e, $\Delta  \subset \Theta$. Hence, if we over-estimate the support of the filter $c$, there will be many linearly independent annihilating filters in the nullspace of $\mathcal{T}(\hat{\rho})$ , or equivalently $\mathcal{T}(\hat{\rho})$ is low rank. 

To avoid oversmoothing of the parameter maps, we choose the spatial support of the 3D filter $\mathbf c$ to be large. This implies that the lifted matrix $\mathbf Q$ is rectangular (more columns than rows) in the parameter mapping setting. 
%
%
\vspace{-1em}

\subsection{Recovery using structured low-rank matrix priors}
In this paper, we focus on the recovery of $T_2$ weighted MR images from their undersampled multichannel encoded measurements, denoted by 
\begin{equation}
\label{measurement_model-compact}
\mathbf{b} = \mathcal{A}({\mathbf{\hat{\rho}}}) + \mathbf{\eta},
\end{equation}
where $\mathcal{A}$ is a linear operator representing Fourier under sampling and multiplication of coil sensitivities with $\hat{\mathbf{\rho}}$. We pose the recovery of the signal \eqref{eq:signal model} as the structured low rank matrix recovery problem: 
\begin{equation}
\label{eq:Schattennorm}
{\mathbf{\hat{\rho}}}^\star = \arg\min_{{\mathbf{\hat{\rho}}}} \|\mathcal{T}(\mathbf{\hat{\rho}})\|_p + \frac{\lambda}{2} \|\mathcal{A}(\mathbf{{\hat{\rho}}}) - \mathbf{b}\|^2_2
\end{equation} 
where $\lambda$ is a regularization parameter and $\|\mathbf{X}\|_p$ is the Schatten $p$ norm, defined as $\|\mathbf{X}\|_p : = \frac{1}{p}{\rm Tr}[(\mathbf{X}^*\mathbf{X})^{\frac{p}{2}}] = \frac{1}{p}{\rm Tr}[(\mathbf{X} \mathbf{X}^*)^{\frac{p}{2}}]=\frac{1}{p}\sum_i \sigma_i^p$; $\sigma_i$ are the singular values of $\mathbf{X}$. Here, $\mathcal{T}(\mathbf{\hat{\rho}})$ denotes the structured multifold Toeplitz matrix, whose entries are the samples of $\hat\rho$; $\hat \rho$ corresponds to the 2-D Fourier coefficients of $\rho$.

\begin{figure}[t!]
\centering
\includegraphics[width=0.49\textwidth]{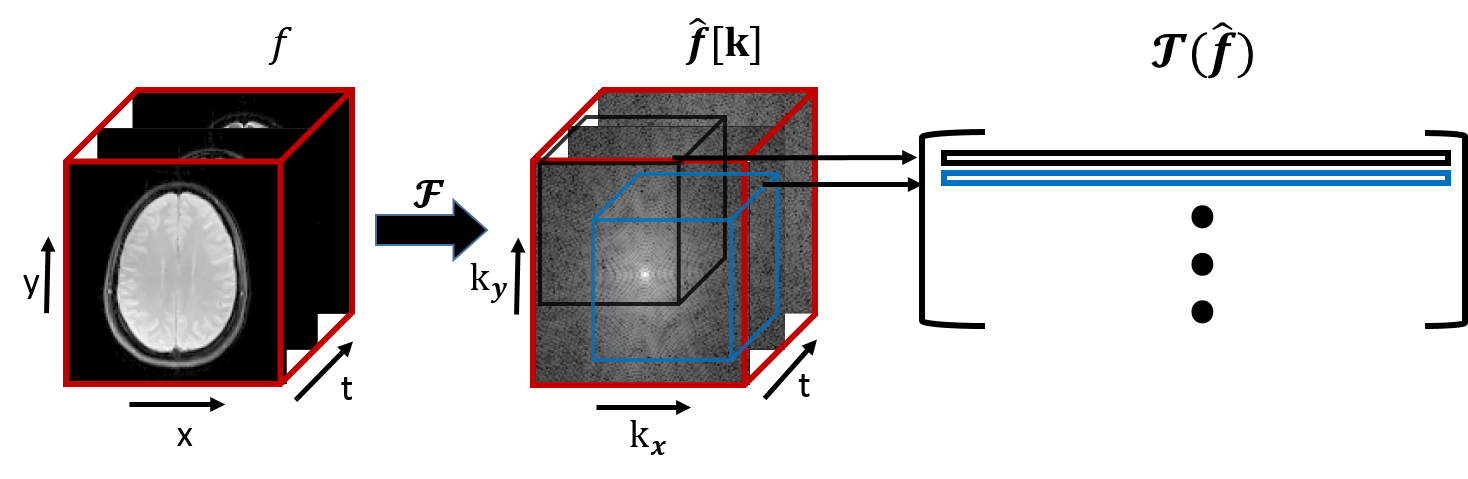}\vspace{-1em}
\caption{Construction of the Toeplitz matrix $\mathcal{T}(\hat{\rho})$: The rows of the matrix correspond to the cube shaped neighborhoods of the Fourier samples. The missing entries of the matrix are estimated by exploiting its low rank structure. }\vspace{-1em}
\label{Fig3}
\end{figure}

\section{Optimization Algorithm}
We rely on the iterative re-weighted least squares (IRLS) based algorithm to solve (\ref{eq:Schattennorm}). The basic idea is to use the identity $\|\mathbf{Y}\|_p =\|\mathbf{H}^{\frac{1}{2}}\mathbf{Y}\|^{2}_F$, where $\mathbf{H} = (\mathbf{Y} \mathbf{Y}^*)^{\frac{p}{2}-1}$. We use an alternating minimization algorithm, which alternates between the following two sub-problems,  to solve \eqref{eq:Schattennorm}:
\begin{alignat}{1}
\label{eq:weight-update}
\mathbf{H}_n & = [\underbrace{\mathcal{T}(\mathbf{\hat{\rho}}_{n-1})\mathcal{T}(\mathbf{\hat{\rho}}_{n-1})^*}_{\mathbf{R}} + \epsilon_n \mathbf{I}]^{\frac{p}{2}-1}\\
\label{eq:x-update}
{\mathbf{\hat{\rho}}}_n & = \arg \min_{{\mathbf{\hat{\rho}}}}\frac{1}{2} \| \mathbf{H}_{n}^{\frac{1}{2}}\mathcal{T}(\mathbf{\hat{\rho}})\|_F^2 + \frac{\lambda}{2}\|\mathcal A(\mathbf{\hat{\rho}}) -\mathbf{b}\|_2^2
\end{alignat}
where $\epsilon_{n} \rightarrow 0$ is added to stabilize the inverse. We will now focus on an efficient implementation of the subproblems.

\vspace{-0.5em}
\subsection{Least squares-Update}
Let the rows of $\mathbf{H}^{\frac{1}{2}}$ be denoted by $\left[(\mathbf{h}^{(1)})^{T},\ldots,(\mathbf{h}^{(M)})^{T}\right]^{T}$. Substituting for $\mathbf{H}^{\frac{1}{2}}$ in \eqref{eq:x-update}, we obtain
\begin{equation}
\label{eq:ls_update}
{\mathbf{\hat{\rho}}}^* = \arg \min_{{\mathbf{\hat{\rho}}}}\frac{1}{2}\sum_{i=1}^{M}\|\mathbf{h}^{(i)}\mathcal{T}(\mathbf{\hat{\rho}})\|^{2}_{2}+\frac{\lambda}{2}\|\mathcal A(\hat{\mathbf{\rho}}) -\mathbf{b}\|_2^2
\end{equation}
Note that the term $\mathbf{h}^{(i)}\mathcal{T}(\mathbf{\hat{\rho}})$ is the linear convolution between the 3-D sequences $\mathbf h^{(i)}$ and $\hat \rho$. In the GIRAF algorithm \cite{GIRAF}, we relied on the approximation of the linear convolution by circular convolutions to accelerate its computations using FFT. We exploited the fast decay of the Fourier coefficients in GIRAF, which made the approximations valid. Here we rely on the annihilation relations in the $k-t$ domain, where the signal does not decay rapidly in the temporal dimension. Hence we use a hybrid approach that consists of performing linear convolutions along time and circular convolutions along the spatial dimension. 

Let $\mathbf{h}^{(i)}$ and $\mathbf{\hat{\rho}}$ consist of $N_{t}$ and $T$ frames respectively and let $k$ denote  $T-N_{t}+1$. Let each frame of the filter $\mathbf{h}^{(i)}$ be of dimension $N_1 \times N_2$. Now $(\mathbf{h}^{(i)} \mathcal{T}(\mathbf{\hat{\rho}}))$ can be simplified as,
\begin{equation}
\label{eq:expand first term}
\mathbf{h}^{(i)} \mathcal{T}(\mathbf{\hat{\rho}}) = \begin{pmatrix}
\mathbf h_{N_{t}}^{(i)}& \ldots & \mathbf h_{1}^{(i)}
\end{pmatrix}\begin{pmatrix}
\mathbf{T}(\mathbf{\hat{\rho}}_{k})& \ldots & \mathbf{T}(\mathbf{\hat{\rho}}_{1})\\
\vdots & \vdots & \vdots \\
\mathbf{T}(\mathbf{\hat{\rho}}_{k+N_{t}-1}) & \cdots & \mathbf{T}(\mathbf{\hat{\rho}}_{N_{t}})\\
\end{pmatrix}
\end{equation}
In the above equation, $\mathbf{T}(\mathbf{\hat{\rho}}_{j})$ represents a Toeplitz matrix formed from $\mathbf{\hat{\rho}}_j$; the product $\mathbf h_i^{(i)}\mathbf{T}(\boldsymbol{\hat{\rho}}_j)$ corresponds to the 2-D convolution between the $i^{th}$ frame of the filter $\mathbf{h}^{(i)}$ and $j^{th}$ frame of $\hat{\rho}$. We safely approximate each of the 2-D convolutions by circular convolutions as in GIRAF and compute them efficiently using Fast Fourier transforms. 

\vspace{-0.5em}
\subsection{Weight-Update}
We consider the Gram matrix $\mathbf{R}$ in \eqref{eq:weight-update} that has the following structure:
\begin{equation}
\label{eq:gram matrix}
\begin{pmatrix}
\mathbf{R}_{1,1} & \mathbf{R}_{1,2} & \ldots & \mathbf{R}_{1,N_{t}}  \\
\vdots & \vdots & \cdots & \vdots  \\
\mathbf{R}_{N_{t},1} & \mathbf{R}_{N_{t},2} & \cdots & \mathbf{R}_{N_{t},N_{t}}  \\
\end{pmatrix}
\end{equation}
with $N_{t}$ column and row partitions and $\mathbf{R}_{i,j}$ is a matrix block of dimension $N_{1} N_{2} \times N_{1} N_{2}$. We obtain the general expression for the matrix block $\mathbf{R}_{p,q}$ as.
\vspace{-0.6em}
\begin{equation}
\label{eq:block-expression for R}
\mathbf{R}_{p,q} = \sum_{i=1}^{k}\mathbf{T}(\mathbf{\hat{\rho}}_{p+i-1}) \mathbf{T}(\mathbf{\hat{\rho}}_{q+i-1})^*
\end{equation}
\vspace{-0.6em}

We observe that the Toeplitz matrix $\mathbf{T}(\hat{\rho}_i)$ can be expressed as $
\mathbf{T}(\mathbf{\hat{\rho}}_i) = \mathbf{P}_{\Lambda}^* \mathsf{Circ}(\mathbf{\hat{\rho}}_{i})\mathbf{P}_{\Gamma}$, 
where $\mathbf{P}_{\Gamma} \in \mathbb{C}^{L \times PQ}$ is a matrix that restricts the convolution onto a valid index set represented by $\Gamma$, $\mathsf{Circ}(\mathbf{\hat{\rho}}_{i}) \in \mathbb{C}^{L \times L}$ is a circulant matrix formed from $\mathbf{\hat{\rho}}_i$ and $\mathbf{P}_{\Lambda}^* \in \mathbb{C}^{N_{1}N_{2} \times L}$ is a matrix representing zero-padding outside the filter support $\Lambda$; here $P, Q$ are the spatial dimensions of $\hat{\mathbf{\rho}}$. Now $\mathbf{T}(\hat{\rho}_i)\mathbf{T}(\hat{\rho}_j)^*$ can be simplified as
\begin{equation}
\label{eq:P_ij modified}
\mathbf{P}_{i,j} = \mathbf{P}_{\Lambda}^*\underbrace{\mathsf{Circ}(\mathbf{\hat{\rho}}_{i})\mathsf{Circ}(\mathbf{\hat{\rho}}_{j})^*}_{\mathsf{Circ}(\mathbf{g})}\mathbf{P}_{\Lambda}
\end{equation}
where the entries of $\mathsf{Circ}(\mathbf{g})$ are generated from the array $g$ given by $\mathbf{F}(\mathbf{F}^{H}(\mathbf{\hat{\rho}}_i)\circ \mbox{conj}(\mathbf{F}^{H}(\mathbf{\hat{\rho}}_j)))$, $\mbox{conj}$ denotes the conjugate operation and $\circ$ denotes point-wise multiplication. 

We evaluate the weight matrix $\mathbf{H}$ as    
\[
\mathbf{H} = [\mathbf{U} (\mathbf{\Lambda}+\epsilon \mathbf{I})\mathbf{ U}^*]^{\frac{p}{2}-1} = \mathbf{U} (\mathbf{\Lambda}+\epsilon \mathbf{ I})^{{\frac{p}{2}-1}} \mathbf{U}^*,
\] 
where $ \mathbf{U}\Lambda\mathbf{U}^*$ is the eigen decomposition of $\mathbf R$. Hence, one choice of the matrix square root $\mathbf{H}^{\frac{1}{2}}$ is $
(\mathbf{\Lambda}+\epsilon \mathbf{I})^{\frac{p}{4}-\frac{1}{2}}\mathbf{U}^*$.


\vspace{-1em}
\section{Experiments}
A fully sampled axial 2-D dataset was acquired on a Siemens 3T Trio scanner using 12 coils and a turbo spin echo sequence. The following scan parameters were used in the acquisition: Matrix size:128$\times$128, FOV: 22$\times$22 $\mbox{cm}^{2}$, TR = 2500 ms and slice thickness = 5 mm. The $T_2$ weighted images were obtained for 12 equispaced echo (TE) times ranging from 10 to 120 ms. Post reconstruction, the $T_2$ maps were obtained by fitting a mono-exponential model to each voxel.

\begin{figure}[t!]
\centering
\includegraphics[width=0.49\textwidth]{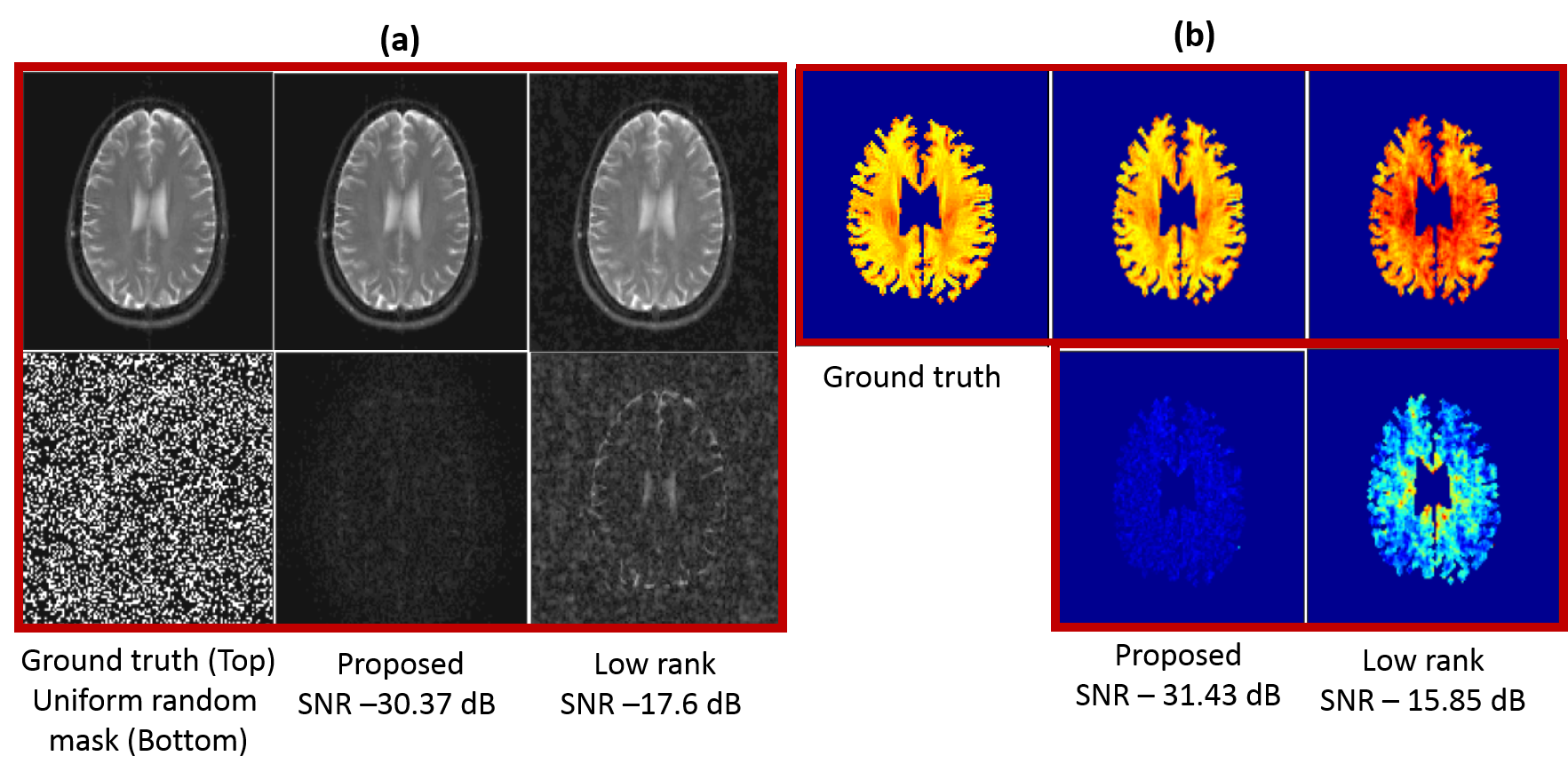}\vspace{-1em}
\caption{Comparison of the proposed method with $k-t$ low rank on the recovery of single channel data from 30 percent uniform random measurements. The improvements offered by the proposed scheme can be easily appreciated from the estimated $T_2$ error images in (a) and the $T_2$ estimates in (b).}\vspace{-1.2em}
\label{Fig5}
\end{figure}

We first study the effect of the filter size (dimensions of the block Toeplitz matrix) on the SNR of the images recovered from 8-fold undersampled multichannel Fourier data in Table. \ref{tab:table1}. We observe that the spatial dimensions of the filter have the largest influence on the SNR, as seen from Table. \ref{tab:table1}.(b). Specifically, we observe that filters with smaller support provides improved results, which demonstrates the benefit of exploiting spatial smoothness; a filter with size $128\times128\times 10$ fails to exploit smoothness. We also observe from the Table.\ref{tab:table1}.(a) the benefit of using temporal annihilation relations. Specifically, we obtain a 0.5 dB improvement over the filter with size 122x122x1, which only exploits joint sparsity, by increasing the length of the filter along time.

In Fig. \ref{Fig5}, we compare the proposed approach with $k-t$ low rank algorithm on the recovery of single coil (coil compressed) $T_2$ weighted data from $30\%$ uniform random measurements. We chose a filter of size $122\times122\times2$ and a Schatten $p=0.6$. We observe that the proposed scheme provides lower errors (see caption for details).

\begin{figure}[t!]
\centering
\includegraphics[width=0.47\textwidth]{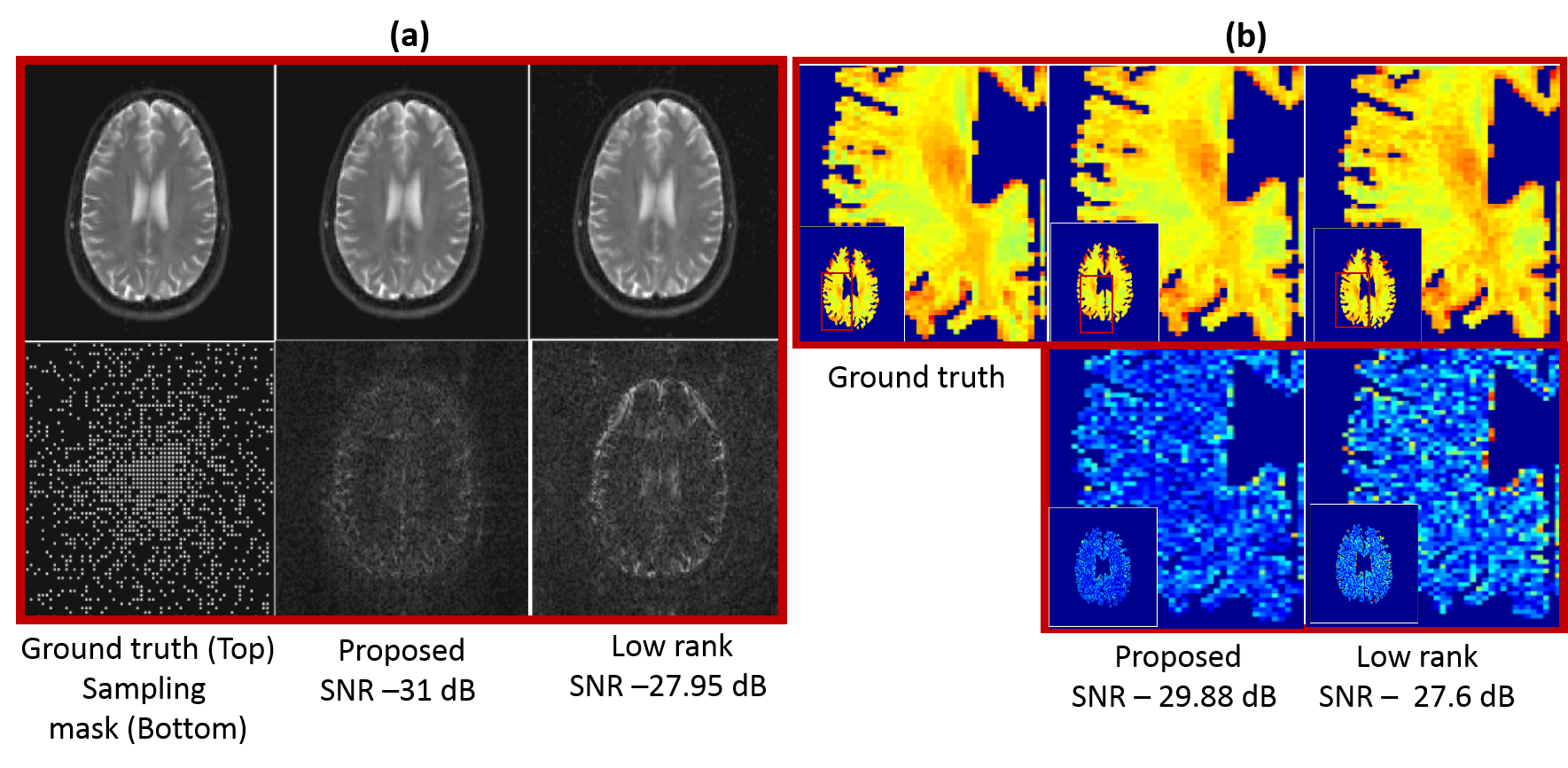}\vspace{-1em}
\caption{Comparison of the proposed method with $k-t$ low rank on the recovery of multi channel data at an acceleration of 12. We observe that the reconstructions from the proposed method have fewer errors, which can be appreciated from the error maps of the $T_2$ weighted images in (a) as well with the noise-like artifacts in the $T_2$ maps in (b).}\vspace{-1.2em}
\label{Fig6}
\end{figure}

In Fig. \ref{Fig6}, we compare the two methods on the recovery of multi-channel $T_2$ weighted data from 12-fold (3-fold variable density + 4-fold 2$\times$2 Cartesian) undersampled data.  We chose a filter of size $114\times114\times10$ and a Schatten $p=0.7$ for the proposed scheme (see caption for details). 
%

\vspace{-0.5em}
\begin{table}[htbp]
  \small
  \centering
  \caption{Effect of filter size on SNR of $T_2$ weighted images.}
  \subfloat[Varying temporal dimension]{%
    \hspace{.2cm}%
    \begin{tabular}{c|r}
        \hline
        filter size       & SNR (dB)  \\
        \hline
        122x122x11       & 31.80  \\
        \hline
        122x122x10       & 32.1  \\
        \hline
        122x122x7       & 32  \\
        \hline
        122x122x2       & 31.84  \\
        \hline
        122x122x1       & 31.56  \\
        \hline
    \end{tabular}%
    \hspace{.2cm}%
  }\hspace{0,2cm}
  \subfloat[Varying spatial dimensions]{%
    \hspace{.2cm}%
       \begin{tabular}{c|r}
        \hline
        filter size       & SNR (dB)  \\
        \hline
        128x128x10       & 29.9  \\
        \hline
        122x122x10       & 32.1  \\
        \hline
        118x118x10       & 32.44  \\
        \hline
        116x116x10       & 32.49  \\
        \hline
        114x114x10       & 32.54  \\
        \hline
    \end{tabular}%
    \hspace{.2cm}%
  }
  \label{tab:table1}
\end{table}\vspace{-0.5em}

%
\vspace{-1em}
\section{Conclusion}
We introduced a novel structured low rank algorithm to recover a 3-D volume consisting of a linear combination of exponentials, from undersampled Fourier measurements. A convolution relation was obtained between the $k-t$ Fourier samples and a 3-D FIR filter by exploiting the exponential structure of the time series at every pixel and the smoothness of the parameters. To speed up the computations, a hybrid approach was employed which consisted of performing circular and linear convolutions along the spatial and temporal dimensions respectively. The algorithm was applied in the context of MR parameter mapping and the reconstructed images and maps were sharper and had fewer errors than those obtained from state of the art methods.
\vspace{-1em}

\bibliographystyle{IEEEtran}

\bibliography{ref_trunc}






\end{document}